\documentclass[letterpaper]{article} 
\usepackage{aaai2026}
\usepackage{times}
\usepackage{helvet}
\usepackage{courier}
\usepackage[hyphens]{url}
\usepackage{graphicx}
\usepackage{natbib}
\usepackage{caption}
\title{Grounded Evaluation and Repair for NL-to-PDDL Problem Generation}
\author{
	Joana Rosa\textsuperscript{\rm 1,3},
	Pedro Santos\textsuperscript{\rm 1},
	Valdemar Oliveira\textsuperscript{\rm 4},
	Rom\~ao Silva\textsuperscript{\rm 4},
	L. Miguel Silveira\textsuperscript{\rm1, \rm 2, \rm 3},
	Bruno Martins\textsuperscript{\rm 2, \rm 3}
}
\affiliations{
	\textsuperscript{\rm 1}INESC INOV, Lisbon, Portugal\\
	\textsuperscript{\rm 2}INESC ID, Lisbon, Portugal\\
	\textsuperscript{\rm 3}Instituto Superior T\'ecnico, Universidade de Lisboa, Lisbon, Portugal\\
	\textsuperscript{\rm 4}Motamineral Minerais Industriais S.A.\\
	\{joana.rosa, pedro.santos\}@inov.pt\\
	\{bruno.g.martins, lms\}@tecnico.ulisboa.pt\\
	\{valdemar.oliveira, romao.silva\}@mota-sc.com
}

\begin{document}

\maketitle

\begin{abstract}
Large Language Models (LLMs) have shown promise for translating Natural Language (NL) planning descriptions into PDDL problem instances. However, standard evaluation criteria such as syntactic validity or planner success can substantially overestimate faithfulness to the described task: a generated problem may be parseable and solvable while misrepresenting the intended initial state, goal, object structure, or optimization target. This paper studies an end-to-end NL-to-PDDL pipeline that combines LLM generation, checks in terms of PDDL parsing, planning and validation, a domain-conformance checker, an LLM critic, and iterative repair. Fine-grained repair feedback is constructed from the domain description, the generated problem, the natural language problem description, and operational diagnostics. 
Reference-based comparisons against curated benchmark PDDL problem descriptions are used for post-hoc benchmark analysis, and these offline checks include renaming-invariant structural matching and semantic equivalence, where domain support is available. Across Planetarium, AutoPlanBench, and curated PDDL~2.1 problems, results show that operational success and benchmark-reference reconstruction can diverge substantially. Results also show that structured repair can be useful, and that PDDL~2.1 remains challenging for reference reconstruction, even when operational success improves.
\end{abstract}

\section{Introduction}
Large Language Models (LLMs) are increasingly being used to translate Natural Language (NL) task descriptions into formal planning representations that can be solved by symbolic planners \cite{liu2023llmp, gestrin2024nl2plan}. Within this setting, NL-to-PDDL generation has emerged as a central problem. The Plan Domain Definition Language (PDDL) is the standard formal language for symbolic planning, being supported by many classical and temporal planners. Still, writing PDDL specifications requires substantial expertise. Translating natural language task descriptions into PDDL offers a natural interface between non-expert users and symbolic planning systems, but it is also demanding: small errors in object declarations, initial predicates, goals, optimization metrics, or numeric fluents, may significantly alter the resulting planning problem.

A major difficulty in this area concerns evaluation. A generated instance may be parseable and even solvable, while encoding the wrong initial state, omitting required predicates, specifying a subtly incorrect goal, or optimizing the wrong quantity. Stronger result validation protocols are therefore needed beyond parseability and planner success alone \cite{zuo2024planetarium}, although strong semantic evaluation is challenging and likely only available for a restricted subset of planning domains.

Considering the aforementioned challenges, this paper reports a detailed analysis guided by three main questions: how can NL-to-PDDL generation be evaluated in a deployment-realistic setting that considers few-shot examples and iterative repairs; how large is the gap between operational acceptance and stricter reference-based benchmark reconstruction metrics; and how well do operational evaluation and repair procedures transfer to PDDL~2.1 benchmarks with temporal and numeric constructs. Through experiments, we show that few-shot prompting and iterative repair improve operational performance in several settings, although reconstructing the curated benchmark references remains difficult, especially in the case of PDDL~2.1.

Feedback-guided repair is an established strategy in LLM-assisted planning model generation, where prior work has used formal, symbolic, validator, and environment feedback to refine generated models. This paper studies the same general repair philosophy in the narrower setting of NL-to-PDDL generation of problem descriptions, assuming that the domain model is already given. Concretely, we combine parser, domain-conformance, planner, validation with VAL, and NL critic feedback in a unified problem generation loop. We also distinguish operational acceptance from benchmark-reference reconstruction, and empirically examine how these signals diverge across Planetarium, AutoPlanBench, and curated PDDL~2.1 settings.

\section{Background}

Research connecting LLMs and automated planning has developed along two broad directions. One line of work studies whether language models can act as planners directly \cite{verma2025teaching}, often generating or selecting plans from NL task descriptions \cite{valmeekam2023planbench,silver2024generalized}. Another direction instead uses language models as planning formalizers, translating user descriptions into structured representations that can be processed by symbolic planning systems \cite{oswald2026modelspace}. This second direction has become increasingly important because it separates two difficult problems: interpreting natural language and solving the resulting planning problem once a correct formal model is available.

Planning formalization has indeed been increasingly treated as a language generation task. Some studies focus on restricted forms of formalization, such as translating NL goals into structured planning goals \cite{xie2023nlgoals}, while others consider richer settings in which complete planning representations must be recovered from text \cite{huang2024limit}. Recent datasets have pushed the problem beyond closed benchmark descriptions and toward open-domain procedural text, showing that performance degrades substantially once the input becomes less templated and more semantically demanding \cite{zhang2024proc2pddl}. A related line of work has also begun to address domain generation, moving beyond problem-instance specification toward full planning model acquisition from natural language \cite{gestrin2024nl2plan,oswald2024domain}.

Several closely related systems use feedback to refine generated planning models. Guan et al. use PDDL validators and human corrective feedback to improve generated world/domain models. In turn, Mahdavi et al. use environment-interaction feedback for automated PDDL translation and planning, while Oswald et al. study symbolic feedback-driven search over planning domain model spaces, including validator output \cite{guan2023worldmodels,mahdavi2024environment,oswald2024domain,oswald2026modelspace}. Recent surveys position this line of work as part of a broader shift from using LLMs directly as planners toward using them as planning formalizers that construct or refine symbolic planning models for downstream planners \cite{tantakoun2025formalizers}. Together, these studies show that formal, symbolic, and interaction-based feedback are increasingly central to LLM-assisted planning formalization.

Related work has further shown that language models can be combined with formal verification or satisfiability-based reasoning tools to handle planning problems more reliably than by direct plan generation alone \cite{hao2024formal}.

\section{Method}

\begin{figure}[t]
	\centering
	\includegraphics[width=0.98\columnwidth]{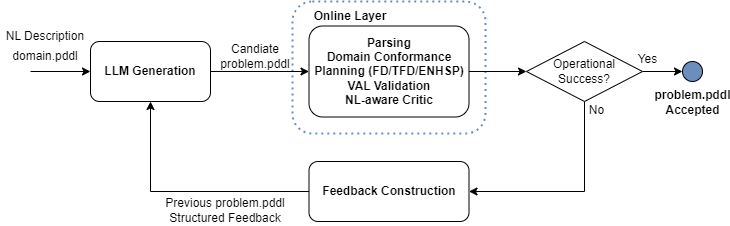}
	\caption{Overview of the proposed NL-to-PDDL generation, evaluation, and repair pipeline.}
	\label{fig:pipeline}
\end{figure}

We define a pipeline that takes as input a NL problem description and a PDDL domain file and produces a candidate \texttt{problem.pddl}. The overall workflow combines initial generation, operational evaluation, feedback construction, and iterative repair. As shown in Figure~\ref{fig:pipeline}, the same generation step is reused throughout the loop: a candidate instance is generated, checked for operational acceptability, and either accepted or revised through structured feedback. The same architecture is used for both classical PDDL and PDDL~2.1, with the latter additionally requiring metric-sensitive checking and planner selection that distinguishes temporal-only from numeric-fluent problems.

\subsection{Initial Generation}

Given a NL problem description and a PDDL domain file, a language model is prompted to generate a complete \texttt{problem.pddl} instance, including objects, initial state, goals, and, when applicable, optimization metrics and numeric initializations. The output must be consistent with the predicate vocabulary and action schema defined in the PDDL problem domain specification.

In selected settings, the prompt is augmented with a small number of same-domain few-shot examples. Each example pairs a NL planning description with its corresponding reference \texttt{problem.pddl}, thereby demonstrating how descriptions in that domain map to object declarations, initial facts, and goal conditions. These examples are distinct from the target instance and serve only as demonstrations of the intended translation pattern.

The model must infer the correct inventory of objects, determine which relations belong in the initial state, identify the intended goal, recover the intended optimization target when one is present, and preserve consistency with the formal domain definition.

\subsection{Online Operational Evaluation and Offline Benchmark Analysis}

The pipeline distinguishes between \emph{online operational evaluation}, which determines whether the repair loop should stop, and \emph{offline benchmark analysis}, which is computed for benchmarking and when a reference \texttt{problem.pddl} is available. This distinction is central to the experimental methodology adopted in this work.

\paragraph{Online Operational Evaluation.}
Each generated problem is first subjected to a sequence of checks.
\begin{enumerate}
	\item \textbf{Parsing:} The generated \texttt{problem.pddl} description must be parseable.
	\item \textbf{Domain Conformance:} A static checker verifies that all referenced objects are declared, predicate names belong to the domain vocabulary, predicate arities are respected, and typing constraints are not violated. The checker is intentionally local and conservative, in the sense that it detects violations of the given domain rather than proving full semantic faithfulness to the NL description.
	\item \textbf{Planning and Validation:} If parsing and static checks succeed, a planner is executed and any returned plan is validated with VAL\footnote{\url{https://github.com/KCL-Planning/VAL}}. For classical domains, we use the Fast Downward planner\footnote{\url{https://www.fast-downward.org/latest/}}. For PDDL~2.1, time-simple domains are handled with TFD\footnote{\url{https://tfd.informatik.uni-freiburg.de/}}, and numeric-fluent domains are processed with ENHSP\footnote{\url{https://sites.google.com/view/enhsp/}}.
	\item \textbf{LLM Critic:} An additional LLM critic compares the NL description with the generated \texttt{problem.pddl}, taking into account the domain file and the automated diagnostics. When it rejects an input instance, it returns a structured judgment identifying each detected problem, its location, and its type.
\end{enumerate}

A candidate is accepted only if it is parseable, domain-conformant, solvable, VAL-valid, and accepted by the LLM critic. We refer to this conjunction of conditions as \emph{operational success}. This criterion is deliberately stricter than parser/planner success alone, yet it does not depend on access to a curated reference problem file.

The LLM critic is instructed to be conservative. In particular, it must anchor its claims in explicit evidence from the NL description, the domain, or the generated PDDL. It must also distinguish object-existence issues from state-level issues, and prefer reporting ambiguity rather than inventing unsupported mismatches. Together, parse success, domain conformance, planner success, VAL validity, and critic acceptance, provide the operational backbone of the pipeline.

\paragraph{Stopping Criteria.}
The repair loop stops under any of the following conditions.
\begin{itemize}
	\item \textbf{Operational Success:} All the aforementioned online checks succeed.
	\item \textbf{Unchanged Generation:} The current \texttt{problem.pddl} is identical to the previous attempt, indicating that the model is not making progress under the current feedback.
	\item \textbf{No Actionable Feedback:} The evaluator cannot provide concrete repair guidance, even though operational success has not been reached.
	\item \textbf{Maximum Attempts Exhausted:} The configured value for the maximum number of attempts is reached.
\end{itemize}

The \textit{unchanged generation} and \textit{no actionable feedback} conditions are important in practice because they prevent the loop from wasting attempts once it has become stagnant or diagnostically uninformative.

\paragraph{Semantic Evaluation:}
When the domain has semantic support, the generated problem can be evaluated against the reference instance through a semantic-equivalence procedure. In this work, this is possible for domains supported by the Planetarium \cite{zuo2024planetarium} benchmark, which represents PDDL problems as graphs over objects and propositions, and checks equivalence after completing partially specified goals. This support is domain-specific, as semantic evaluation is available only for domains for which the required oracle and graph-construction functions have been implemented. Note that semantic equivalence is used only as an \emph{offline benchmark-analysis signal}. This signal is highly informative for reference-based analysis, but it is not used inside the repair loop because a reference problem instance would not be available.

\paragraph{Structural Evaluation.}
In all runs, a renaming-invariant structural comparison can be computed between the generated problem and the reference instance. The comparison checks four components explicitly:
\begin{itemize}
	\item the typed object declarations in \texttt{:objects},
	\item the atomic facts in \texttt{:init},
	\item the atomic facts in \texttt{:goal},
	\item the optimization objective in \texttt{:metric}.
\end{itemize}

For PDDL~2.1, the same comparison also evaluates numeric fluents structurally, by checking the corresponding assignments and numeric initial values in canonical form. The metric clause is evaluated strictly: both the optimization direction (e.g., \texttt{minimize} vs.\ \texttt{maximize}) and the optimized expression itself must match after canonicalization.

Rather than requiring literal identity of object names, the comparison abstracts away from naming and searches for a consistent bijection between generated and reference objects. To reduce unnecessary computation, the structural comparison first checks whether an identity mapping already yields a match when the object names coincide. Only if that fast path fails does it search over alternative bijections. After aligning objects under the selected mapping, it compares the resulting object declarations, initial facts, goal facts, and metric clause. The comparison is order-insensitive and based on a canonical PDDL problem representation (e.g., line comments are ignored before fact extraction so that annotations do not introduce spurious mismatches).

When semantic support is available, both semantic equivalence and structural matching can be computed offline. However, neither is used as the online acceptance criterion, due to the dependence on access to a curated reference problem file. Instead, both serve as post-hoc measures of benchmark-reference reconstruction, allowing us to quantify the gap between operational success and recovery of the curated reference encoding. An example for the structured matching procedure is provided in Appendix~\ref{app:structural-example}.

\subsection{Iterative Repair}

Whenever the generated problem fails the online acceptance criterion, feedback is constructed and returned to the model for repair. The feedback includes both coarse-grained and fine-grained signals. Coarse-grained feedback reports whether the instance parsed successfully, whether it passed the domain checker, whether a plan was found, whether validation succeeded, and whether the critic accepted it. Fine-grained feedback is constructed from the checker diagnostics, planner/validator outcomes, and critic output, identifying issues such as undeclared objects, predicate-arity mismatches, domain-incompatible facts, critic-identified initialization or goal mismatches, or suspected discrepancies between the NL description and the generated problem.

As illustrated in Figure~\ref{fig:pipeline}, the repair loop does not rely on a separate repair module. Instead, the generation step is invoked again with the same task inputs, augmented with the previously generated \texttt{problem.pddl} and the structured feedback derived from evaluation. The loop continues until operational success is reached, the generation stops changing, no actionable feedback is available, or the maximum attempt budget is exhausted.

\subsection{Offline Benchmark Analysis Across Different Planning Tasks}

The same generate-and-repair pipeline was applied across both the Planetarium and the AutoPlanBench benchmarks. When reference problems are available, we compute benchmark-based analysis signals offline. In domains with Planetarium semantic support, this includes both semantic equivalence and renaming-invariant structural matching. In unsupported domains, the available offline signal is the structural criterion alone. This yields a unified operational pipeline with broader domain coverage and stronger post-hoc analysis whenever benchmark references exist.

We additionally considered PDDL~2.1 planning, using six manually curated benchmark domains derived from IPC 2002 domains: \texttt{depot-numeric}, \texttt{depot-time-simple}, \texttt{driverlog-numeric}, \texttt{driverlog-time-simple}, \texttt{rovers-numeric}, and \texttt{rovers-time-simple}. In this setting, the generation and repair logic are the same, while planner support is split across Temporal Fast Downward (TFD) for time-simple domains, and the Expressive Numeric Heuristic Search Planner (ENHSP) for numeric-fluent domains. This allows testing the pipeline on temporal and numeric PDDL~2.1 settings, using efficient planners according to different problem requirements. Overall, we assessed a unified operational evaluation-and-repair framework for NL-to-PDDL tasks, that can be applied across benchmarks and planning formalisms, while retaining semantic and structural comparison as offline benchmark analysis signals whenever they are available.

\section{Experimental Setup}

We now discuss the considered experimental setup.

\subsection{Benchmarks and Data}

As mentioned in the previous section, two main benchmark sources were used for classical planning experiments. The first was Planetarium \cite{zuo2024planetarium}, which provides native support for semantic equivalence checking in selected domains. The second was AutoPlanBench \cite{stein2025autoplanbench}, which provides NL resources across a broader range of planning domains.

AutoPlanBench was integrated through a dataset adapter that normalizes each instance into a common representation, consisting of a NL input, a domain file, and a reference \texttt{problem.pddl} file. In this setting, the NL input was not read from per-instance description files. Instead, it was generated automatically from each reference \texttt{problem.pddl} using AutoPlanBench's domain-level natural language resources, which provide object-name mappings and predicate verbalizations through templates. This yields a standardized NL description of the objects, initial state, and goal, while preserving the benchmark's domain-specific verbalization.

The experimental analysis covers three main subsets.

\begin{itemize}
	\item \textbf{Classical, Semantically Supported:} Planetarium and AutoPlanBench domains for which Planetarium-style semantic support is available, namely \texttt{blocksworld}, \texttt{gripper} and \texttt{floor-tile}/\texttt{floortile}. Each run attempts 60 problem instances.
	\item \textbf{Classical, Structurally Evaluated:} AutoPlanBench domains without Planetarium semantic support, namely \texttt{depot}, \texttt{logistics} and \texttt{satellite}. A total of 60 problem instances are considered.
	\item \textbf{PDDL~2.1, Structurally Evaluated:} Six sets of manually curated domain variants derived from IPC 2002, namely \texttt{depot-numeric}, \texttt{depot-time-simple}, \texttt{driverlog-numeric}, \texttt{driverlog-time-simple}, \texttt{rovers-numeric}, and \texttt{rovers-time-simple}. Each of the six sets contains ten benchmark instances together with a same-domain few-shot example.
\end{itemize}

To characterize the benchmark instances more concretely, we also computed simple size statistics from the curated reference problems. Classical semantically supported examples contain on average 8.2 objects (range 4--19), 13.0 initial facts (6--57), and 4.8 goal atoms (1--11). Unsupported classical AutoPlanBench examples are somewhat larger in the initial state, with 13.4 objects (6--18), 20.0 initial facts (5--28), and 2.5 goal atoms (1--5). The curated PDDL~2.1 examples contain 9.1 objects (7--13), 23.4 initial facts or numeric initializations (10--44), and 1.6 goal atoms (1--3), with a metric clause present in all instances.

\subsection{Prompting Conditions}

Experiments were conducted with \texttt{gpt-4.1-mini} as the base generation model and the critic. Each example was processed with up to three attempts, consisting of one initial generation followed by up to two repair iterations. All LLM calls used a default decoding temperature of \(0.0\), including generation, repair, and critic calls.

Two prompting conditions were considered:
\begin{itemize}
	\item \textbf{Baseline:} Direct generation from the PDDL domain file and natural language description.
	\item \textbf{Baseline + Few-Shot:} The baseline prompt augmented with one same-domain few-shot example.
\end{itemize}

Few-shot selection is dataset-specific: Planetarium uses local same-domain examples with a compatibility fallback when needed, while AutoPlanBench uses the benchmark's domain-specific few-shot resources aligned with the adapted \texttt{problem.pddl}. PDDL~2.1 uses one same-domain example from each of the six benchmark sets.

The initial prompt asks the model to generate only a valid \texttt{problem.pddl} file from the natural language description, target domain name, and exact domain PDDL, including objects, initial state, goals, and metrics when applicable. The repair prompt reuses the same inputs, adds the previous draft and structured evaluator feedback, and asks for the smallest set of edits needed to restore domain compliance and faithfulness to the NL description. The critic prompt is separated from both generation and repair: it receives the NL description, domain, generated problem, and automated diagnostics, and returns a structured acceptance judgment together with localized issues when the instance should be revised.
Additional runs with \texttt{gpt-5.5} and \texttt{opus-4.7} are reported in Appendix~\ref{app:results-with-different-llms} and show the same qualitative pattern: operational acceptance is often high, while reference-based reconstruction remains more variable, especially in PDDL~2.1.

\subsection{Evaluation Protocol}

All planning-based checks were performed with the Fast Downward solver and with VAL in classical domains. In the case of PDDL~2.1 problems, time-simple domains were evaluated with TFD and VAL, while numeric-fluent domains were evaluated with ENHSP and VAL.

For all experiments, the online pipeline used the same operational acceptance criterion: a candidate \texttt{problem.pddl} was accepted only if it parsed, passed the domain-conformance checker, yielded a plan, produced a VAL-valid plan, and was accepted by the LLM critic. If any of these conditions failed, structured feedback was constructed and returned to the model together with the previous candidate, yielding an iterative repair loop that executes for up to three total attempts.

When benchmark references were available, renaming-invariant structural matching was computed offline after the run, with semantic equivalence available as an additional signal in domains supported by the corresponding Planetarium oracle. These offline metrics quantify how often an operationally accepted instance also reconstructs the curated benchmark reference problem. In other words, they measure the gap between deployment-realistic acceptance and benchmark-reference reconstruction.

\subsection{Evaluation Metrics}

The evaluation protocol separates operational acceptance from benchmark-reference fidelity. Offline structural and semantic comparisons should not be interpreted as assessing whether a generated PDDL problem is the only correct representation of the intended planning instance. Rather, they measure how closely the generated problem reconstructs the curated benchmark reference encoding, up to the equivalences supported by each benchmark. For this reason, the analysis combines operational acceptance metrics, offline benchmark analysis metrics, and repair-oriented measures.

The main online metric is \emph{operational success rate} (Op.), which corresponds to the fraction of examples that satisfy the full acceptance criterion:
\[
\mathrm{OpSucc} = \frac{1}{N} \sum_{i=1}^{N}
\mathbf{1}[\mathrm{parse}_i \wedge \mathrm{check}_i \wedge \mathrm{solve}_i \wedge \mathrm{VAL}_i \wedge \mathrm{critic}_i].
\]
We also report the individual component rates in appendix: parse success, domain conformance (Domain Conf.), solve success, VAL plan validity (VAL-valid), and critic acceptance (Critic Acc.). Metrics are reported for the initial generation (Step-0) and/or for the final output after any repair. Gains denote final minus Step-0 accuracy for the corresponding metric. Each experimental run evaluates results over 60 examples, unless stated otherwise due to timeouts. The term \(N\) denotes completed examples, so \(60-N\) corresponds to example-level timeouts. We also report average end-to-end runtime per completed example.

Because the loop may terminate without success, we additionally report the main non-success stopping outcomes: unchanged generation (Stop: Unchanged) and maximum-attempt exhaustion (Stop: Max Att.). The no-actionable-feedback condition did not occur in the reported runs and is omitted from the tables. These statistics are important for interpreting whether repair failures arise from stagnation or from limits in the attempt budget.

Using the benchmark references, we additionally compute offline reference-based metrics. Renaming-invariant structural accuracy (Struct.) is the common offline signal across all benchmarks, defined as exact agreement with the curated reference problem up to a consistent object renaming over \texttt{:objects}, \texttt{:init}, \texttt{:goal}, and \texttt{:metric}. This criterion is intentionally strict, in that a failed structural match does not necessarily imply that the generated problem is invalid or useless as a planning instance. Instead, it indicates that the generated problem does not recover the particular reference encoding used by the benchmark. We therefore interpret structural accuracy as benchmark-reference reconstruction accuracy, and not as an absolute test assessing the preservation of task semantics.
In semantically supported domains, semantic equivalence (Sem.) is also computed through the corresponding domain-specific oracle. We explicitly report the divergence between online and offline criteria through \emph{Op. Non-Struct.}, i.e. the fraction of examples accepted operationally but not structurally matched, and \emph{Struct. Non-Op.}, i.e. the fraction of structurally matched but not operationally accepted instances. The offline reference-based metrics can be written under a common formulation:
\[
\mathrm{EqAcc} = \frac{1}{N} \sum_{i=1}^{N} \mathbf{1}[\hat{p}_i \equiv p_i].
\]
In the previous equation, $\hat{p}_i$ is the generated problem, $p_i$ is the curated reference problem, and $\equiv$ denotes the chosen equivalence relation, i.e. structural or semantic equivalence depending on the available benchmark support.

To quantify iterative improvement in the offline analysis, \emph{repair gain} is defined as the difference between the final reference-match rate and the reference-match rate at the initial generation step:
\[
\mathrm{RepairGain} = \mathrm{RefMatch}_{\mathrm{final}} - \mathrm{RefMatch}_{\mathrm{step0}}.
\]

Bold rows in the result tables denote gain metrics, computed as the difference between the final attempt and Step-0 for the corresponding measure.

For all reported settings, we compare operational success with structural benchmark matching in order to identify cases in which an instance is accepted by the online pipeline but fails the offline reference-match criterion, or matches the reference structurally but fails an operational check.

\section{Experimental Results}

We first analyze classical domains with semantic support, then unsupported classical domains, and finally PDDL~2.1 domains with temporal and numeric constructs. Throughout the different tests, operational metrics determine online stopping, while semantic and structural metrics are computed offline when supported.

\subsection{Classical Planning with Semantic Support}

Domains for which semantic support is available are considered first. Table~\ref{tab:classical-semantic-main} compares final operational acceptance with offline semantic and structural reference matching for the Planetarium (PL) and AutoPlanBench (APB) benchmarks.
Table~\ref{tab:classical-semantic-main} compares the operational outcomes with offline reference-based metrics. Few-shot prompting improves both semantic and structural reference matching on Planetarium, with final semantic equivalence rising from 0.267 to 0.417 and final structural matches from 0.200 to 0.417. AutoPlanBench performs substantially better overall, but the few-shot run is lower than the baseline under both offline criteria, with final semantic and structural matches decreasing from 0.842 to 0.714. The gain rows show that offline reference-match gains are smaller than operational gains: semantic gain is positive only on Planetarium, and structural gain is at most 0.017. The divergence rows further show that operational acceptance and benchmark-reference reconstruction remain distinct signals.

A domain-level inspection of the same run shows uneven behavior: \texttt{blocksworld} reaches 0.900 final operational success, and \texttt{floor-tile} reaches 0.600, although \texttt{gripper} collapses to 0.000. This shows that few-shot prompting is not merely refining already-strong outputs: for some domains it is the difference between a viable operational result and near-complete failure.
A closer inspection suggests that this weakness is dataset-specific rather than inherent to the \texttt{gripper} domain alone. In the Planetarium baseline run, \texttt{gripper} reaches 0.000 final operational success, with failures split between unchanged generations and maximum-attempt exhaustion. The most frequent critic issues concern initial-state bookkeeping and goal interpretation, especially ball-location facts, carried-ball facts, and free-gripper facts. With few-shot prompting, Planetarium \texttt{gripper} improves to 0.500 final operational success and 0.850 VAL-validity. By contrast, AutoPlanBench \texttt{gripper} is stronger, reaching 0.842--0.789 operational success and 0.947--1.000 structural reference matching across the two prompting conditions. This suggests that the failures arise from the interaction between \texttt{gripper} descriptions, initial-state conventions, and the critic/repair loop, rather than from the domain vocabulary alone.

\begin{table}[hbp!]
		\centering
		\small
		\begin{tabular}{lcccc}
		\hline
		Metric & Pl. Base & Pl. +FS & APB Base & APB +FS \\
		\hline
		Step-0 Op. & 0.333 & 0.517 & 0.649 & 0.661 \\
		Final Op. & 0.500 & 0.583 & 0.772 & 0.750 \\
		\textbf{Op. Gain} & \textbf{0.167} & \textbf{0.067} & \textbf{0.123} & \textbf{0.089} \\
		\hline
		Step-0 Sem. & 0.183 & 0.400 & 0.842 & 0.714 \\
		Final Sem. & 0.267 & 0.417 & 0.842 & 0.714 \\
		\textbf{Sem. Gain} & \textbf{0.083} & \textbf{0.017} & \textbf{0.000} & \textbf{0.000} \\
		\hline
		Step-0 Struct. & 0.183 & 0.400 & 0.842 & 0.714 \\
		Final Struct. & 0.200 & 0.417 & 0.842 & 0.714 \\
		\textbf{Struct. Gain} & \textbf{0.017} & \textbf{0.017} & \textbf{0.000} & \textbf{0.000} \\
		\hline
		Op. Non-Struct. & 0.300 & 0.200 & 0.123 & 0.196 \\
		Struct. Non-Op. & 0.000 & 0.033 & 0.193 & 0.161 \\
		\hline
	\end{tabular}
	\caption{Operational acceptance (Op.) versus semantic and structural reference matching (Sem./Struct.) on semantically supported classical problem domains.}
	\label{tab:classical-semantic-main}
\end{table}

Detailed operational component checks are reported in Table~\ref{tab:classical-semantic-diagnostics}, in Appendix~\ref{app:results-with-different-llms}. All reported columns use the operational stopping criterion. Semantic and structural matching are reported as offline benchmark metrics.

The detailed diagnostics in Appendix~\ref{app:results-with-different-llms}, specifically in Table~\ref{tab:classical-semantic-diagnostics}, show that few-shot prompting improves operational success on Planetarium, raising the final rate from 0.500 to 0.583, but slightly lowers it on APB, from 0.772 to 0.750. Operational success is also stricter than parser/planner success alone, because solve and VAL-valid rates are higher than final operational success once the critic acceptance is included. Iterative feedback improves operational success in all four settings, with larger gains in the baseline runs.

\subsection{Classical Planning without Semantic Support}

The pipeline was next evaluated on unsupported AutoPlanBench domains, namely \texttt{depot}, \texttt{logistics}, and \texttt{satellite}. In this setting, semantic equivalence is unavailable, so renaming-invariant structural matching is the offline benchmark signal.
Table~\ref{tab:classical-structural-main} compares operational acceptance with structural reference matching. Few-shot prompting improves final structural exact match from 0.567 to 0.650, but structural gain is negative in both conditions (i.e., -0.100 and -0.067), showing that the repair loop can move candidates away from exact benchmark reconstruction. The divergence rows show that few examples are accepted operationally while failing structural matching (0.033 and 0.017). The larger divergence is in the opposite direction: 0.183 of examples in both conditions match structurally but are not operationally accepted.

\begin{table}[hbp!]
	\centering
	\small
	\begin{tabular}{lcc}
		\hline
		Metric & APB Base & APB +FS \\
		\hline
		Step-0 Op. & 0.400 & 0.483 \\
		Final Op. & 0.417 & 0.483 \\
		\textbf{Op. Gain} & \textbf{0.017} & \textbf{0.000} \\
		\hline
		Step-0 Struct. & 0.667 & 0.717 \\
		Final Struct. & 0.567 & 0.650 \\
		\textbf{Struct. Gain} & \textbf{-0.100} & \textbf{-0.067} \\
		\hline
		Op. Non-Struct. & 0.033 & 0.017 \\
		Struct. Non-Op. & 0.183 & 0.183 \\
		\hline
	\end{tabular}
	\caption{Operational acceptance (Op.) versus structural reference matching (Struct.) on unsupported AutoPlanBench problem domains.}
	\label{tab:classical-structural-main}
\end{table}

Table~\ref{tab:classical-structural-diagnostics} in Appendix~\ref{app:results-with-different-llms} reports the operational run outcomes.
The detailed diagnostics in Table~\ref{tab:classical-structural-diagnostics} show that few-shot prompting improves final operational success from 0.417 to 0.483 and reduces the two main non-success stopping outcomes. However, operational gains are small in the baseline run and zero with few-shot prompting, indicating that most of the few-shot improvement comes from a stronger initial generation rather than from repair.

\subsection{Comparing Operational Acceptance Versus Reference Matching in Classical Runs}

The main divergence in the classical runs exists between operational success and benchmark-reference reconstruction. 
Planetarium+FS accepts 0.583 of examples operationally, but only 0.417 match the reference semantically and structurally, yielding 0.200 operationally accepted but structurally unmatched cases. Supported APB reaches higher benchmark-reference reconstruction, with final semantic and structural accuracy of 0.842 in the baseline condition and 0.714 with few-shot prompting. However this setting also contains structurally matched examples that are not operationally accepted. Unsupported APB shows the same latter pattern: final structural matching is higher than final operational success in both baseline and few-shot settings.

\subsection{Extension to PDDL~2.1}

Evaluation was finally extended to PDDL~2.1 domains covering both time-simple and numeric variants. The goal of this extension was not merely to verify end-to-end executability, but also to test whether the same operational loop is effective once temporal planning and numeric fluents are introduced. Table~\ref{tab:pddl21-main} shows that the PDDL~2.1 outputs often become operationally executable even when they do not recover the curated reference instance. Final structural exact match remains very low, rising only from 0.000 to 0.067, and structural gain is small (0.000 and 0.033). The divergence rows make the gap explicit: operationally accepted but structurally unmatched outputs account for 0.500 of the baseline run and 0.667 of the few-shot run, while no structurally matched output fails operational acceptance. This indicates that operational feedback helps produce accepted PDDL~2.1 instances but does not by itself recover metric-sensitive benchmark structure.

\begin{table}[htb!]
	\centering
	\small
	\begin{tabular}{lcc}
		\hline
		Metric & Baseline & Baseline + FS \\
		\hline
		Step-0 Op. & 0.350 & 0.567 \\
		Final Op. & 0.500 & 0.733 \\
		\textbf{Op. Gain} & \textbf{0.150} & \textbf{0.167} \\
		\hline
		Step-0 Struct. & 0.000 & 0.034 \\
		Final Struct. & 0.000 & 0.067 \\
		\textbf{Struct. Gain} & \textbf{0.000} & \textbf{0.033} \\
		\hline
		Op. Non-Struct. & 0.500 & 0.667 \\
		Struct. Non-Op. & 0.000 & 0.000 \\
		\hline
	\end{tabular}
	\caption{Operational acceptance (Op.) versus structural reference matching (Struct.) on the six curated PDDL~2.1 problem domain variants.}
	\label{tab:pddl21-main}
\end{table}

Detailed operational component checks, reported in Table~\ref{tab:pddl21-diagnostics} within Appendix~\ref{app:results-with-different-llms}, show that few-shot prompting improves the operational side substantially: final operational success rises from 0.500 to 0.733, solve and VAL-valid rates rise from 0.550 to 0.750, and critic acceptance rises from 0.517 to 0.767. Few-shot prompting also reduces maximum-attempt exhaustion from 0.417 to 0.217.

\subsection{Manual Audit of the LLM Critic}

To sanity-check the LLM critic, we manually inspected 24 stratified outputs, covering classic and PDDL~2.1 problems. The assessment criterion was faithfulness to the natural language description, not exact reconstruction of the curated reference file. The critic agreed with the manual judgment in 13/24 cases: it accepted 9 manually faithful and 3 manually unfaithful instances, and rejected 8 manually faithful and 4 manually unfaithful instances. These results support using the critic as an operational feedback signal, but not as an independently reliable task-faithfulness oracle. The full results are provided in Appendix~\ref{app:manual-audit-examples}, namely the audit statistics in Table~\ref{tab:critic-manual-audit} and a few representative examples.

\section{Discussion}
Across semantically supported settings, planner-oriented operational metrics are not interchangeable with reference-based benchmark reconstruction. The pipeline makes this explicit by using operational success as the online stopping criterion and measuring semantic and structural reference matching offline, whenever a benchmark reference is available. The Planetarium+FS run illustrates the gap clearly: operational acceptance improves from 0.517 at step~0 to 0.583 after repair, yet final semantic and structural reference-match rates reach only 0.417. APB results are stronger overall, with final semantic and structural reference-match rates between 0.714 and 0.842 on completed examples.
The results also show that operational acceptance and benchmark-reference reconstruction are separate signals. Operational success is necessary for deployment, but benchmark-reference reconstruction and per-domain robustness remain important for controlled scientific evaluation. Overall, the task of reconstructing PDDL problems from natural language descriptions remains particularly challenging.

Another main finding concerns few-shot prompting, which is beneficial but not uniformly across benchmarks. On Planetarium it improves both operational and offline reference-based results. On unsupported APB it raises final operational success from 0.417 to 0.483, and final structural matching from 0.567 to 0.650. On PDDL~2.1 problems, it raises final operational success from 0.500 to 0.733. However, on supported APB instances, the few-shot run is slightly lower than the baseline on both operational and offline reference-based metrics. This suggests that few-shot examples are useful but not uniformly positive. Iterative repair also varies by setting: it contributes more under the operational metric than under the structural metric, especially when the feedback can make an executable problem acceptable without recovering the exact benchmark instance.

More broadly, the same generate-and-repair framework can be applied across benchmarks and formalisms, but any conclusions should consider the evaluation signal that is available. In classical APB, structural matching is often recovered. In PDDL~2.1, operational success is much easier to obtain than exact structural agreement, suggesting that temporal and numeric problem generation requires stronger feedback about numeric initializations, metric clauses, and curated reference object structure. Progress in NL-to-PDDL generation therefore depends not only on stronger generators, but also on stronger operational critics, stronger benchmark analysis, and planner-aware diagnostics.
The main error patterns involve incomplete initial states, structurally plausible reference mismatches, representation-sensitive object/predicate/metric mismatches, and PDDL~2.1-specific failures involving omitted metrics or incomplete numeric initializations. Additional details about error patterns are provided in Appendix~\ref{app:error-patterns}.

\section{Limitations and Future Work}

The present study has several important limitations. First, the online acceptance criterion is only a proxy for task faithfulness. The domain checker is intentionally local, the planner and validator only assess executability, and the LLM critic can produce false positive or false negative validations. As a result, operational success should not be interpreted as a proof that the generated instance fully captures the intended task.
Our pipeline validates plans for the generated problem with VAL, but it does not yet compare the behavior induced by the generated and reference instances. Future work should complement reference-file reconstruction with plan-oriented checks, such as testing plan transfer between generated and reference problems, or comparing valid, optimal, or top-quality plan sets.

A second limitation concerns the type of natural language descriptions used in the experiments. Much of the benchmark data is relatively formal and close to the structure that is ultimately required in the target PDDL instance, even in the manually curated PDDL~2.1 domains. The PDDL~2.1 benchmark design introduces more contextual variation than direct benchmark verbalization through templates, but it falls short of genuinely open-ended user language. A broader evaluation across different levels of linguistic informality, underspecification, and common-sense inference, would therefore be a natural next step.

A third limitation concerns the fact that the pipeline assumes the availability of a correct \texttt{domain.pddl} file and focuses only on the generation of \texttt{problem.pddl} descriptions. This makes the setup suitable for controlled benchmarking, but it leaves aside the more difficult task of generating the domain model itself. Recent work has begun to address this broader problem, including approaches that generate both domain and problem specifications from natural language descriptions \cite{gestrin2024nl2plan}, and work focused specifically on domain generation \cite{oswald2024domain,oswald2026modelspace}. Extending the present framework in that direction would considerably broaden its scope.

A further limitation concerns the LLM critic. Although it can flag mismatches that parsers, planners, and validators cannot observe, it remains an LLM judgment and may reject faithful encodings or accept plausible but incomplete ones. Our small manual audit confirms this limitation: the critic agreed with manual faithfulness judgments in 13/24 inspected cases, with both false positives and false negatives. Larger manual studies are still needed to measure critic reliability across domains, prompt variants, and error types, and to determine when the LLM critic feedback indeed improves subsequent repair attempts.

The present study is also restricted to a single-turn interaction setting, in which the user provides the complete problem description at once. This leaves open a more interactive scenario in which the generated PDDL specification is explained back to the user, revised through follow-up dialogue, or incrementally corrected after clarification. Such interaction would be especially valuable when descriptions are informal, incomplete, or ambiguous.

Finally, the extension to PDDL~2.1 should be viewed as an initial step rather than a comprehensive evaluation. This setup covers six curated domain variants spanning time-simple and numeric formulations, but it does not yet establish performance on richer temporal benchmarks, more expressive numeric formulations, or broader planner coverage. The low structural exact-match rate in PDDL~2.1 also shows that operational checks and critic feedback are not sufficient to recover metric-sensitive benchmark instances reliably. Extending the framework to more demanding PDDL~2.1/2.2 domains, alternative numeric planners, and eventually to probabilistic settings through PPDDL, is another important direction for future work.

\section{Conclusions}

This paper argues that NL-to-PDDL generation cannot be evaluated reliably through parseability or planner success alone. We present a LLM-based generation pipeline that adopts an operationally grounded stopping criterion based on parsing, domain conformance, planning, VAL validation, and an LLM critic, while reserving benchmark comparison for offline analysis. Across Planetarium, AutoPlanBench, and PDDL~2.1 experimental runs, the results show that operationally successful instances may fail stricter reference-based benchmark reconstruction criteria. Semantic equivalence and renaming-invariant structural comparison therefore serve as offline benchmark analysis signals that quantify reconstruction of the curated reference instance.

Our results also clarify the roles of few-shot prompting and iterative repair. Their impact is not uniform across settings, in that few-shot prompting improves Planetarium, unsupported APB, and PDDL~2.1 operational success, but not the supported APB run. Repair effects are most visible when operational feedback can turn an executable near-miss into an accepted instance. At the same time, the unsupported APB and PDDL~2.1 results show that operational repair is not equivalent to structural benchmark recovery.

Deployment-realistic repair should be driven by operational evidence, but controlled benchmarks should continue to measure reconstruction of the curated reference instance separately from operational acceptance. Progress in NL-to-PDDL generation depends not only on stronger generators, but also on stronger critics, stronger benchmark analysis, and repair mechanisms that work across settings.

\section*{Acknowledgements}

This research was developed in the scope of the project ``Pacto de Inovação ECP -- Ecocerâmica e Cristalaria de Portugal'', with reference 01/C05-i11/2024.PC644916391-00000029,  call number 02/C05-i01.01/2022, funded by the Portuguese Recovery and Resilience Program (PPR), The Portuguese Republic, and The European Union (EU) under the framework of the Next Generation EU Program.
The work was also supported by Fundação para a Ciência e a Tecnologia (FCT) under projects UID/50021/2025 (\url{https://doi.org/10.54499/UID/50021/2025}), UID/PRR/50021/2025 (\url{https://doi.org/10.54499/UID/PRR/50021/2025}), UID/6486/2025 (\url{https://doi.org/10.54499/UID/06486/2025}), UID/PRR/6486/2025 (\url{https://doi.org/10.54499/UID/PRR/06486/2025}), and UID/PRR2/06486/2025 (\url{https://doi.org/10.54499/UID/PRR2/06486/2025}).

\bibliography{aaai2026}

\appendix
\section*{Appendices}
\section{Additional Error Pattern Discussion}
\label{app:error-patterns}

The observed errors cluster into four recurring categories. First, many failures involve incomplete or inaccurate initial states, where small omissions or incorrect predicates are enough to break structural recovery or semantic equivalence. Second, some outputs are structurally plausible reference mismatches: they are parseable, solvable, and VAL-valid, but fail the offline semantic or structural reference-match criterion. Third, representation-sensitive mismatches involve object declarations, initialization facts, predicates, or metrics that differ from the curated benchmark encoding despite an otherwise plausible task structure. Finally, PDDL~2.1 introduces metric-specific failures, including omitted or incorrect \texttt{:metric} clauses and incomplete numeric initializations.

\section{Additional GPT-4.1-mini Operational Diagnostics}
This appendix reports detailed operational component checks for the \texttt{gpt-4.1-mini} runs, whose summary results are discussed in the main text. Table~\ref{tab:classical-semantic-diagnostics} reports the semantically supported classical diagnostics. Table~\ref{tab:classical-structural-diagnostics} reports the unsupported AutoPlanBench diagnostics. Finally,Table~\ref{tab:pddl21-diagnostics} reports the PDDL~2.1 diagnostics.

\begin{table}[!hbp]
	\centering
		\footnotesize
		\begin{tabular}{lcccc}
		\hline
		Metric & Pl. Base & Pl. +FS & APB Base & APB +FS \\
		\hline
		$N$ & 60 & 60 & 57 & 56 \\
		Step-0 Op. & 0.333 & 0.517 & 0.649 & 0.661 \\
		Final Op. & 0.500 & 0.583 & 0.772 & 0.750 \\
		\textbf{Op. Gain} & \textbf{0.167} & \textbf{0.067} & \textbf{0.123} & \textbf{0.089} \\
		\hline
		Parse & 1.000 & 0.967 & 1.000 & 1.000 \\
		Domain Conf. & 1.000 & 1.000 & 1.000 & 1.000 \\
		Solve & 0.650 & 0.833 & 0.965 & 0.946 \\
		VAL-valid & 0.650 & 0.833 & 0.965 & 0.946 \\
		Critic Acc. & 0.617 & 0.600 & 0.772 & 0.750 \\
		Stop: Unchanged & 0.183 & 0.083 & 0.175 & 0.107 \\
		Stop: Max Att. & 0.317 & 0.333 & 0.053 & 0.143 \\
		Avg. Time (s) & 18.72 & 19.70 & 40.24 & 39.62 \\
			\hline
			\end{tabular}
		\caption{Operational component checks and repair outcomes on semantically supported problem domains.}
			\label{tab:classical-semantic-diagnostics}
\end{table}

\begin{table}[!hbp]
		\centering
		\small
		\begin{tabular}{lcc}
		\hline
		Metric & APB Base & APB +FS \\
		\hline
		$N$ & 60 & 60 \\
		Step-0 Op. & 0.400 & 0.483 \\
		Final Op. & 0.417 & 0.483 \\
		\textbf{Op. Gain} & \textbf{0.017} & \textbf{0.000} \\
		\hline
		Parse & 0.967 & 0.983 \\
		Domain Conf. & 0.983 & 0.967 \\
		Solve & 0.667 & 0.717 \\
		VAL-valid & 0.667 & 0.717 \\
		Critic Acc. & 0.583 & 0.617 \\
		Stop: Unchanged & 0.267 & 0.233 \\
		Stop: Max Att. & 0.317 & 0.283 \\
		Avg. Time (s) & 18.95 & 16.43 \\
		\hline
		\end{tabular}
		\caption{Operational component checks and repair outcomes on unsupported AutoPlanBench problem domains.}
		\label{tab:classical-structural-diagnostics}
\end{table}

\begin{table}[!hbp]
		\centering
		\small
		\begin{tabular}{lcc}
		\hline
		Metric & Baseline & Baseline + FS \\
		\hline
		$N$ & 60 & 60 \\
		Step-0 Op. & 0.350 & 0.567 \\
		Final Op. & 0.500 & 0.733 \\
		\textbf{Op. Gain} & \textbf{0.150} & \textbf{0.167} \\
		\hline
		Parse & 1.000 & 1.000 \\
		Domain Conf. & 0.967 & 0.983 \\
		Solve & 0.550 & 0.750 \\
		VAL-valid & 0.550 & 0.750 \\
		Critic Acc. & 0.517 & 0.767 \\
		Stop: Unchanged & 0.083 & 0.050 \\
		Stop: Max Att. & 0.417 & 0.217 \\
		Avg. Time (s) & 21.77 & 27.31 \\
		\hline
		\end{tabular}
		\caption{Operational component checks and repair outcomes on the six curated PDDL~2.1 problem domain variants.}
		\label{tab:pddl21-diagnostics}
\end{table}

\section{Results with Different LLMs}
\label{app:results-with-different-llms}
This appendix reports experimental results with stronger LLMs.
Although the strongest results are obtained with \texttt{gpt-5.5} and \texttt{opus-4.7}, it remains relevant to examine how more efficient and resource-effective models perform in the context of NL-to-PDDL tasks, and hence our choice of using \texttt{gpt-4.1-mini} in the main results.
Across \texttt{gpt-5.5} and \texttt{opus-4.7}, operational success is near-saturated on most classical settings, while benchmark fidelity remains more variable. The clearest residual gap is in PDDL~2.1, where structural recovery stays low despite high operational acceptance.

\subsection{Results with GPT-5.5}
This appendix reports detailed operational component checks for the \texttt{gpt-5.5} runs.

Table~\ref{tab:appendix-gpt55-classical-semantic-diagnostics} reports operational component checks and repair outcomes, while
Table~\ref{tab:appendix-gpt55-classical-semantic-main} reports a comparison between operational acceptance and semantic and structural benchmark fidelity over the semantically supported domains.
Table~\ref{tab:appendix-gpt55-classical-structural-diagnostics} reports operational component checks and repair outcomes, while
Table~\ref{tab:appendix-gpt55-classical-structural-main} reports a comparison between operational acceptance and semantic and structural benchmark fidelity on unsupported AutoPlanBench problem domains.
Finally, Table~\ref{tab:appendix-gpt55-pddl21-diagnostics} reports operational component checks and repair outcomes, while
Table~\ref{tab:appendix-gpt55-pddl21-main} reports a comparison between operational acceptance and semantic and structural benchmark fidelity on the six curated PDDL~2.1 domain variants.

\begin{center}
	\centering
	\small
	\begin{tabular}{lcccc}
		\hline
		Metric & Pl. Base & Pl. +FS & APB Base & APB +FS \\
		\hline
		$N$ & 59 & 58 & 58 & 57 \\
		Avg. Time (s) & 16.94 & 17.29 & 57.15 & 64.31 \\
		\hline
		Step-0 Op. & 0.966 & 0.931 & 0.983 & 1.000 \\
		Final Op. & 1.000 & 0.983 & 0.983 & 1.000 \\
		\textbf{Op. Gain} & \textbf{0.034} & \textbf{0.052} & \textbf{0.000} & \textbf{0.000} \\
		\hline
		Parse & 1.000 & 1.000 & 0.983 & 1.000 \\
		Domain Conf. & 1.000 & 1.000 & 0.983 & 1.000 \\
		Solve & 1.000 & 1.000 & 1.000 & 1.000 \\
		VAL-valid & 1.000 & 1.000 & 1.000 & 1.000 \\
		Critic Acc. & 1.000 & 0.983 & 1.000 & 1.000 \\
		Stop: Unchanged & 0.000 & 0.017 & 0.000 & 0.000 \\
		Stop: Max Att. & 0.000 & 0.000 & 0.000 & 0.000 \\
		\hline
	\end{tabular}
	\captionof{table}{Operational component checks and repair outcomes on semantically supported problem domains, for \texttt{gpt-5.5}.}
	\label{tab:appendix-gpt55-classical-semantic-diagnostics}
\end{center}

\begin{center}
	\centering
	\small
	\begin{tabular}{lcccc}
		\hline
		Metric & Pl. Base & Pl. +FS & APB Base & APB +FS \\
		\hline
		Step-0 Op. & 0.966 & 0.931 & 0.983 & 1.000 \\
		Final Op. & 1.000 & 0.983 & 0.983 & 1.000 \\
		\textbf{Op. Gain} & \textbf{0.034} & \textbf{0.052} & \textbf{0.000} & \textbf{0.000} \\
		\hline
		Step-0 Sem. & 0.932 & 0.966 & 1.000 & 1.000 \\
		Final Sem. & 0.932 & 0.983 & 1.000 & 1.000 \\
		\textbf{Sem. Gain} & \textbf{0.000} & \textbf{0.017} & \textbf{0.000} & \textbf{0.000} \\
		\hline
		Step-0 Struct. & 0.831 & 0.880 & 0.983 & 1.000 \\
		Final Struct. & 0.814 & 0.914 & 0.983 & 1.000 \\
		\textbf{Struct. Gain} & \textbf{-0.017} & \textbf{0.034} & \textbf{0.000} & \textbf{0.000} \\
		\hline
		Op. Non-Struct. & 0.186 & 0.086 & 0.000 & 0.000 \\
		Struct. Non-Op. & 0.000 & 0.017 & 0.000 & 0.000 \\
		\hline
	\end{tabular}
	\captionof{table}{Operational acceptance (Op.) versus semantic and structural benchmark fidelity (Sem./Struct.) on semantically supported problem domains, for \texttt{gpt-5.5}.}
	\label{tab:appendix-gpt55-classical-semantic-main}
\end{center}

\begin{center}
	\centering
	\small
	\begin{tabular}{lcc}
		\hline
		Metric & APB Base & APB +FS \\
		\hline
		$N$ & 60 & 60 \\
		Avg. Time (s) & 13.08 & 14.94 \\
		\hline
		Step-0 Op. & 0.983 & 0.950 \\
		Final Op. & 1.000 & 1.000 \\
		\textbf{Op. Gain} & \textbf{0.017} & \textbf{0.050} \\
		\hline
		Parse & 1.000 & 1.000 \\
		Domain Conf. & 1.000 & 1.000 \\
		Solve & 1.000 & 1.000 \\
		VAL-valid & 1.000 & 1.000 \\
		Critic Acc. & 1.000 & 1.000 \\
		Stop: Unchanged & 0.000 & 0.000 \\
		Stop: Max Att. & 0.000 & 0.000 \\
		\hline
	\end{tabular}
	\captionof{table}{Operational component checks and repair outcomes on unsupported AutoPlanBench domains, for \texttt{gpt-5.5}.}
	\label{tab:appendix-gpt55-classical-structural-diagnostics}
\end{center}

\begin{center}
	\centering
	\small
	\begin{tabular}{lcc}
		\hline
		Metric & APB Base & APB +FS \\
		\hline
		Step-0 Op. & 0.983 & 0.950 \\
		Final Op. & 1.000 & 1.000 \\
		\textbf{Op. Gain} & \textbf{0.017} & \textbf{0.050} \\
		\hline
		Step-0 Struct. & 0.650 & 0.950 \\
		Final Struct. & 0.667 & 0.983 \\
		\textbf{Struct. Gain} & \textbf{0.017} & \textbf{0.033} \\
		\hline
		Op. Non-Struct. & 0.333 & 0.017 \\
		Struct. Non-Op. & 0.000 & 0.000 \\
		\hline
	\end{tabular}
	\captionof{table}{Operational acceptance (Op.) versus structural benchmark fidelity (Struct.) on unsupported AutoPlanBench problem domains, for \texttt{gpt-5.5}.}
	\label{tab:appendix-gpt55-classical-structural-main}
\end{center}

\begin{center}
	\centering
	\small
	\begin{tabular}{lcc}
		\hline
		Metric & Baseline & Baseline + FS \\
		\hline
		$N$ & 60 & 60 \\
		Avg. Time (s) & 20.70 & 19.85 \\
		\hline
		Step-0 Op. & 0.867 & 0.950 \\
		Final Op. & 0.983 & 0.967 \\
		\textbf{Op. Gain} & \textbf{0.117} & \textbf{0.017} \\
		\hline
		Parse & 1.000 & 1.000 \\
		Domain Conf. & 1.000 & 1.000 \\
		Solve & 1.000 & 0.967 \\
		VAL-valid & 1.000 & 0.967 \\
		Critic Acc. & 0.983 & 0.967 \\
		Stop: Unchanged & 0.000 & 0.000 \\
		Stop: Max Att. & 0.017 & 0.033 \\
		\hline
	\end{tabular}
	\captionof{table}{Operational component checks and repair outcomes on the six curated PDDL~2.1 problem domain variants, for \texttt{gpt-5.5}.}
	\label{tab:appendix-gpt55-pddl21-diagnostics}
\end{center}

\begin{center}
	\centering
	\small
	\begin{tabular}{lcc}
		\hline
		Metric & Baseline & Baseline + FS \\
		\hline
		Step-0 Op. & 0.867 & 0.950 \\
		Final Op. & 0.983 & 0.967 \\
		\textbf{Op. Gain} & \textbf{0.117} & \textbf{0.017} \\
		\hline
		Step-0 Struct. & 0.067 & 0.133 \\
		Final Struct. & 0.067 & 0.133 \\
		\textbf{Struct. Gain} & \textbf{0.000} & \textbf{0.000} \\
		\hline
		Op. Non-Struct. & 0.917 & 0.833 \\
		Struct. Non-Op. & 0.000 & 0.000 \\
		\hline
	\end{tabular}
	\captionof{table}{Operational acceptance (Op.) versus structural benchmark fidelity (Struct.) on the six curated PDDL~2.1 problem domain variants, for \texttt{gpt-5.5}.}
	\label{tab:appendix-gpt55-pddl21-main}
\end{center}

\subsection{Results with Opus 4.7}
This appendix reports detailed operational component checks for the \texttt{opus-4.7} runs.

Table~\ref{tab:appendix-opus47-classical-semantic-diagnostics} reports operational component checks and repair outcomes, while
Table~\ref{tab:appendix-opus47-classical-semantic-main} reports a comparison between operational acceptance and semantic and structural benchmark fidelity over the semantically supported domains.
Table~\ref{tab:appendix-opus47-classical-structural-diagnostics} reports operational component checks and repair outcomes, while
Table~\ref{tab:appendix-opus47-classical-structural-main} reports a comparison between operational acceptance and semantic and structural benchmark fidelity on unsupported AutoPlanBench problem domains.
Finally, Table~\ref{tab:appendix-opus47-pddl21-diagnostics} reports operational component checks and repair outcomes, while
Table~\ref{tab:appendix-opus47-pddl21-main} reports a comparison between operational acceptance and semantic and structural benchmark fidelity on the six curated PDDL~2.1 domain variants.

\begin{center}
	\centering
	\small
	\begin{tabular}{lcccc}
		\hline
		Metric & Pl. Base & Pl. +FS & APB Base & APB +FS \\
		\hline
		$N$ & 60 & 60 & 60 & 60 \\
		Avg. Time (s) & 50.30 & 80.67 & 283.72 & 367.73 \\
		\hline
		Step-0 Op. & 0.967 & 0.917 & 0.950 & 0.967 \\
		Final Op. & 1.000 & 1.000 & 0.983 & 1.000 \\
		\textbf{Op. Gain} & \textbf{0.033} & \textbf{0.083} & \textbf{0.033} & \textbf{0.033} \\
		\hline
		Parse & 1.000 & 1.000 & 1.000 & 1.000 \\
		Domain Conf. & 1.000 & 1.000 & 0.983 & 1.000 \\
		Solve & 1.000 & 1.000 & 1.000 & 1.000 \\
		VAL-valid & 1.000 & 1.000 & 1.000 & 1.000 \\
		Critic Acc. & 1.000 & 1.000 & 1.000 & 1.000 \\
		Stop: Unchanged & 0.000 & 0.000 & 0.000 & 0.000 \\
		Stop: Max Att. & 0.000 & 0.000 & 0.000 & 0.000 \\
		\hline
	\end{tabular}
	\captionof{table}{Operational component checks and repair outcomes on semantically supported domains, for \texttt{opus-4.7}.}
	\label{tab:appendix-opus47-classical-semantic-diagnostics}
\end{center}

\begin{center}
	\centering
	\small
	\begin{tabular}{lcccc}
		\hline
		Metric & Pl. Base & Pl. +FS & APB Base & APB +FS \\
		\hline
		Step-0 Op. & 0.967 & 0.917 & 0.950 & 0.967 \\
		Final Op. & 1.000 & 1.000 & 0.983 & 1.000 \\
		\textbf{Op. Gain} & \textbf{0.033} & \textbf{0.083} & \textbf{0.033} & \textbf{0.033} \\
		\hline
		Step-0 Sem. & 0.967 & 0.983 & 0.983 & 1.000 \\
		Final Sem. & 1.000 & 1.000 & 1.000 & 1.000 \\
		\textbf{Sem. Gain} & \textbf{0.033} & \textbf{0.017} & \textbf{0.017} & \textbf{0.000} \\
		\hline
		Step-0 Struct. & 0.734 & 0.950 & 0.983 & 1.000 \\
		Final Struct. & 0.767 & 0.967 & 1.000 & 1.000 \\
		\textbf{Struct. Gain} & \textbf{0.033} & \textbf{0.017} & \textbf{0.017} & \textbf{0.000} \\
		\hline
		Op. Non-Struct. & 0.233 & 0.033 & 0.000 & 0.000 \\
		Struct. Non-Op. & 0.000 & 0.000 & 0.017 & 0.000 \\
		\hline
	\end{tabular}
	\captionof{table}{Operational acceptance (Op.) versus semantic and structural benchmark fidelity (Sem./Struct.) on semantically supported problem domains, for \texttt{opus-4.7}.}
	\label{tab:appendix-opus47-classical-semantic-main}
\end{center}

\begin{center}
	\centering
	\small
	\begin{tabular}{lcc}
		\hline
		Metric & APB Base & APB +FS \\
		\hline
		$N$ & 59 & 60 \\
		Avg. Time (s) & 93.60 & 39.78 \\
		\hline
		Step-0 Op. & 0.881 & 1.000 \\
		Final Op. & 0.932 & 1.000 \\
		\textbf{Op. Gain} & \textbf{0.051} & \textbf{0.000} \\
		\hline
		Parse & 0.932 & 1.000 \\
		Domain Conf. & 0.932 & 1.000 \\
		Solve & 1.000 & 1.000 \\
		VAL-valid & 1.000 & 1.000 \\
		Critic Acc. & 1.000 & 1.000 \\
		Stop: Unchanged & 0.000 & 0.000 \\
		Stop: Max Att. & 0.000 & 0.000 \\
		\hline
	\end{tabular}
	\captionof{table}{Operational component checks and repair outcomes on unsupported AutoPlanBench problem domains, for \texttt{opus-4.7}.}
	\label{tab:appendix-opus47-classical-structural-diagnostics}
\end{center}

\begin{center}
	\centering
	\small
	\begin{tabular}{lcc}
		\hline
		Metric & APB Base & APB +FS \\
		\hline
		Step-0 Op. & 0.881 & 1.000 \\
		Final Op. & 0.932 & 1.000 \\
		\textbf{Op. Gain} & \textbf{0.051} & \textbf{0.000} \\
		\hline
		Step-0 Struct. & 0.593 & 0.667 \\
		Final Struct. & 0.610 & 0.667 \\
		\textbf{Struct. Gain} & \textbf{0.017} & \textbf{0.000} \\
		\hline
		Op. Non-Struct. & 0.322 & 0.333 \\
		Struct. Non-Op. & 0.000 & 0.000 \\
		\hline
	\end{tabular}
	\captionof{table}{Operational acceptance (Op.) versus structural benchmark fidelity (Struct.) on unsupported AutoPlanBench problem domains, for \texttt{opus-4.7}.}
	\label{tab:appendix-opus47-classical-structural-main}
\end{center}

\begin{center}
	\centering
	\small
	\begin{tabular}{lcc}
		\hline
		Metric & Baseline & Baseline + FS \\
		\hline
		$N$ & 60 & 60 \\
		Avg. Time (s) & 98.75 & 51.82 \\
		\hline
		Step-0 Op. & 0.833 & 0.983 \\
		Final Op. & 0.933 & 1.000 \\
		\textbf{Op. Gain} & \textbf{0.100} & \textbf{0.017} \\
		\hline
		Parse & 0.950 & 1.000 \\
		Domain Conf. & 0.950 & 1.000 \\
		Solve & 0.966 & 1.000 \\
		VAL-valid & 0.966 & 1.000 \\
		Critic Acc. & 0.966 & 1.000 \\
		Stop: Unchanged & 0.017 & 0.000 \\
		Stop: Max Att. & 0.017 & 0.000 \\
		\hline
	\end{tabular}
	\captionof{table}{Operational component checks and repair outcomes on the six curated PDDL~2.1 problem domain variants, for \texttt{opus-4.7}.}
	\label{tab:appendix-opus47-pddl21-diagnostics}
\end{center}

\begin{center}
	\centering
	\small
	\begin{tabular}{lcc}
		\hline
		Metric & Baseline & Baseline + FS \\
		\hline
		Step-0 Op. & 0.833 & 0.983 \\
		Final Op. & 0.933 & 1.000 \\
		\textbf{Op. Gain} & \textbf{0.100} & \textbf{0.017} \\
		\hline
		Step-0 Struct. & 0.233 & 0.200 \\
		Final Struct. & 0.250 & 0.217 \\
		\textbf{Struct. Gain} & \textbf{0.017} & \textbf{0.017} \\
		\hline
		Op. Non-Struct. & 0.683 & 0.783 \\
		Struct. Non-Op. & 0.000 & 0.000 \\
		\hline
	\end{tabular}
	\captionof{table}{Operational acceptance (Op.) versus structural benchmark fidelity (Struct.) on the six curated PDDL~2.1 problem domain variants, for \texttt{opus-4.7}.}
	\label{tab:appendix-opus47-pddl21-main}
\end{center}

\section{Prompt Templates}

The generation and repair prompts were shared across \texttt{gpt-4.1-mini}, \texttt{gpt-5.5}, and \texttt{opus-4.7}. However, some small adjustments where made to the LLM critic prompt for the \texttt{gpt-5.5} and \texttt{opus-4.7} runs.

The templates below correspond to the prompts used to define the role, output format, and task-level constraints for each LLM call. In the API call, the system prompt and user prompt are sent as separate messages in the same request: the former provides general instructions, while the latter is a structured JSON payload containing the concrete benchmark instance and any dynamic repair context. 

The generation user prompt has the following structure:
\begin{quote}\scriptsize\ttfamily
	\{\\
	\quad "task": "Generate a PDDL problem from a natural language description for the provided benchmark domain.",\\
	\quad "domain\_name": ...,\\
	\quad "problem\_name\_hint": ...,\\
	\quad "natural\_language\_description": ...,\\
	\quad "domain\_pddl": ...,\\
	\quad "few\_shot\_examples": [...]\\
	\}
\end{quote}

For repair calls, the same instance-specific inputs are reused, and the repair user prompt is augmented with the previous draft, repair history, and evaluator-derived feedback, as shown next:
\begin{quote}\scriptsize\ttfamily
	\{\\
	\quad "task": "Repair the generated PDDL problem using the feedback.",\\
	\quad "domain\_name": ...,\\
	\quad "problem\_name\_hint": ...,\\
	\quad "natural\_language\_description": ...,\\
	\quad "domain\_pddl": ...,\\
	\quad "previous\_problem\_pddl": ...,\\
	\quad "repair\_history": [...],\\
	\quad "few\_shot\_examples": [...],\\
	\quad "feedback": ...\\
	\}
\end{quote}

The few-shot examples are supplied dynamically through the \texttt{few\_shot\_examples} element in the user prompt, for both generation and repair calls. Repair-specific information is supplied dynamically through \texttt{previous\_problem\_pddl}, \texttt{repair\_history}, and \texttt{feedback}, after each evaluation step.

\subsection{Generation Prompt}

Used for \texttt{gpt-4.1-mini}, \texttt{gpt-5.5}, and \texttt{opus-4.7}.

\begin{quote}\scriptsize\ttfamily
	You are an expert PDDL problem generator.\\ \\ Your task is to generate only a PDDL problem file from:\\ - a natural language planning task description.\\ - the target domain name.\\ - the exact domain PDDL.\\ \\ Requirements:\\ - Output only valid PDDL problem text. No markdown and no explanations.\\ - The generated problem must use the provided domain name exactly.\\ - Reuse only predicates, typing rules, constants, and syntax compatible with the provided domain PDDL.\\ - Infer objects, initial state, and goal from the natural language only.\\ - Do not generate extra requirements not implied by the domain/problem description.\\ - Prefer complete and explicit object declarations and explicit initialization and goal facts.\\ - Include all required facts exactly, including 0-arity predicates such as (arm-empty) when implied.\\ - Aim for a problem that is faithful to the natural language task, compatible with the\\   provided domain, and solvable by a planner.\\ - Internally do reasoning in three stages before writing the final answer:\\   1. identify the full object inventory and object types/constants used by the task.\\   2. derive the complete initial state.\\   3. derive the complete goal state.\\ - Before producing the final PDDL problem specification, internally verify that every object mentioned in\\   :init and :goal is declared in :objects, and that every fact is supported by the natural language description plus the provided domain.\\ - When the natural language description is partially underspecified or symmetric, prefer the most\\   literal and canonical interpretation rather than an arbitrary alternative instantiation.\\ - Be conservative: do not omit unary, support, location, holding, availability, or\\   emptiness facts that are required to make the intended state fully specified.\\ - If few-shot examples are provided, use them as domain-specific demonstrations of how\\   natural language maps to a complete problem.pddl specification for this domain.\\ - Learn domain-specific conventions from the few-shot examples, but do not copy object\\   names or instance-specific facts, unless they are supported by the current task.\\ - Ensure balanced parentheses and standard PDDL formatting.
\end{quote}

\subsection{Repair Prompt}

Used for \texttt{gpt-4.1-mini}, \texttt{gpt-5.5}, and \texttt{opus-4.7}.

\begin{quote}\scriptsize\ttfamily
	You are an expert PDDL repair assistant.\\ \\ You will receive:\\ - the natural language task.\\ - the exact domain PDDL.\\ - the current generated PDDL problem draft that must be repaired.\\ - an optional history of previous attempts.\\ - operational feedback from parser, domain, planner, and validator checks.\\ \\ Your job is to return a corrected PDDL problem.\\ \\ Requirements:\\ - Output only valid PDDL problem text. No markdown, no explanations.\\ - Preserve the original natural language meaning.\\ - Use the provided domain name exactly.\\ - Treat the provided previous\_problem\_pddl content as a draft to edit, not as disposable\\   context.\\ - Preserve any parts of the current draft that are already correct.\\ - Apply the smallest set of changes needed to satisfy the feedback.\\ - Repair syntax, objects, initial facts, and goal facts as needed.\\ - Do not change the domain PDDL.\\ - Return a fully corrected problem that is faithful to the natural language task,\\   compatible with the provided domain, and solvable by a planner.\\ - Include any missing 0-arity predicates, such as (arm-empty), whenever they are required by the task.\\ - Be especially careful with initialization and goal facts that differ by only one relation, one\\   support block, or one unary predicate.\\ - Treat the feedback as operational repair guidance:\\   - fix any listed syntax, domain-name, predicate, arity, object declaration, typing, planner, or validator issues.\\   - compare every :init and :goal fact against the provided domain PDDL before returning\\     the repaired problem\\   - do not use predicates, object types, constants, or syntax that are unsupported by\\     the domain.\\ - Prefer minimal edits over full rewrites of :init and :goal facts.\\ - If the current draft already has the correct objects section, keep it unchanged.\\ - After repairing, internally check that the problem parses, conforms to the domain, and\\   gives the planner a coherent initial state and a reachable goal.\\ - If few-shot examples are provided, use them only as domain-specific guidance for the\\   structure of correct problem instances in this domain.\\ - If feedback conflicts with the natural language description, prefer the natural language description plus the domain constraints.
\end{quote}

\subsection{LLM Critic Prompt}

The following critic prompt was used directly for \texttt{gpt-4.1-mini} runs, and some minor adaptations were used for \texttt{gpt-5.5} and \texttt{opus-4.7}.

\begin{quote}\scriptsize\ttfamily
	You are auditing a generated PDDL problem for faithfulness to a natural language\\   planning task.\\ \\ You will receive:\\ - the original natural language task.\\ - the exact domain PDDL.\\ - the automatically generated PDDL problem.\\ - automatic parser/domain/planner/validator diagnostics.\\ - optionally, the plan found for the generated problem.\\ \\ Your job is to compare the generated problem against the natural language task description and the\\   domain.\\ Do not assume access to any ground-truth PDDL problem.\\ Do not rewrite the PDDL.\\ \\ Return only a JSON object with this schema:\\ \{\\ "accepted": true or false,\\ "issues": [\\ \{\\ "category": "nl\_mismatch | init\_mismatch | goal\_mismatch | object\_mismatch |\\         domain\_misuse | underspecified | other",\\ "severity": "high | medium | low",\\ "scope": "objects | init | goal | predicates | types | actions | plan\_interpretation |\\         other",\\ "focus": "short label naming exactly what the issue refers to",\\ "message": "concise explanation"\\ \}\\ ],\\ "repair\_instructions": ["concise instruction", "..."]\\ \}\\ \\ Acceptance criteria:\\ - Accept if the generated objects, initial state, and goal are faithful to the natural language description as far as can be judged from the text and domain.\\ - Reject if the generated problem is formally valid but appears to solve a weaker,\\   different, contradictory, or materially incomplete task.\\ - Base every reported issue on explicit evidence from the natural language task description, the generated PDDL, or the automatic diagnostics. Do not speculate about hidden intent,\\   benchmark conventions, likely plans, or facts that are not stated.\\ - Be conservative about rejection. Do not reject solely because of harmless object\\   renaming, formatting, ordering, redundant facts, or additional final-state facts that\\   are consistent with the requested outcome.\\ - If more than one interpretation is plausible and the generated PDDL problem matches one reasonable literal interpretation of the text, prefer acceptance rather than inventing\\   a mismatch.\\ - Do not reject extra goal facts that are natural consequences of the requested final\\   arrangement or common domain bookkeeping, unless they make the task materially\\   stricter in a way that contradicts the natural language request.\\ - Treat the automatic parser/domain/planner/validator diagnostics as authoritative for\\   formal compatibility. Do not report a domain\_misuse issue if the automatic diagnostics\\   say the problem is parseable, domain-compatible, and plan-valid, unless you can point\\   to a natural language faithfulness issue.\\ - If the automatic diagnostics are all successful, default to acceptance unless you can\\   identify a concrete contradiction or omission relative to the natural language task.\\ - Remember that a PDDL problem specifies only objects, initial state, and goal. Do not\\   reject because intermediate actions or transitions are not explicitly represented in\\   the problem; the planner is responsible for deriving the action sequence.\\ - If the initial state and goal imply that some intermediate action must occur, that is\\   not an issue by itself. Reject only if the initial state or goal contradicts or omits\\   something materially stated in the natural language.\\ - Check cardinality carefully: phrases like "each", "all", "both", "every", "all\\   destinations", or "all endpoints" usually require facts for every mentioned entity,\\   not just one representative.\\ - Respect the domain's predicate and function signatures. Do not suggest a repair that\\   removes required arguments or creates facts that cannot be expressed with the provided\\   domain.\\ - Distinguish object existence from state predicates. If the natural language says a\\   resource/status is not available, empty, free, occupied, loaded, assigned, etc.,\\   represent that through the relevant domain predicates when possible; do not remove\\   typed objects or other domain-required facts unless the text clearly says the objects\\   themselves do not exist.\\ - When natural language and domain constraints interact, propose repairs that satisfy\\   both: preserve objects and facts required for a coherent domain instance while\\   adding/removing only the state facts that the text supports.\\ - Do not suggest deleting or negating facts that appear operationally necessary for\\   domain executability unless the natural language explicitly forbids them and you can\\   name a coherent domain-compatible alternative.\\ - If a fact seems required by the domain but appears in tension with the\\   natural language wording, explicitly report the conflict or ambiguity instead of\\   blindly removing the fact.\\ - Prefer repair instructions such as "clarify whether this refers to object existence or\\   state availability" or "preserve domain-required facts and revise only the conflicting\\   state interpretation" over instructions that would make the problem operationally\\   incoherent.\\ - Stay consistent across your own analysis:\\   - do not first acknowledge that a fact is domain-required or compatible with the\\     domain and later recommend removing it without explaining what domain-compatible\\     replacement will make the problem executable.\\   - do not alternate between "this fact should be present" and "this fact should be\\     removed" unless you explicitly identify the ambiguity that caused the change in\\     judgment.\\   - if the issue is genuinely ambiguous, say so directly and keep the repair instruction\\     conservative.\\ - Prefer stable, minimal repair guidance over oscillating advice. If one interpretation\\   preserves domain executability and another breaks it, prefer the executable interpretation unless the natural language clearly rules it out.\\ - If you reject, make the highest-severity issues actionable and tied to a concrete\\   contradiction or omission in objects, init, or goal. Avoid labeling an issue as high\\   severity while explaining that the generated PDDL is actually consistent.\\ - For every issue, make the target of the complaint explicit:\\   - use "scope" to say whether the issue is about objects, :init, :goal,\\     predicate/function usage, types, action interpretation, or something else.\\   - use "focus" to name the exact entity or concept at stake, such as "color objects",\\     "robot-has facts", "available-color facts", "goal cardinality", or "tile adjacency".\\   - if the issue is about state availability rather than object existence, say that\\     explicitly instead of implying that the objects should be removed.\\ - Do not hide the target only inside prose. The structured fields must make clear what\\   the issue refers to even if the message is read quickly.\\ - Keep the verdict internally consistent:\\   - if you conclude the generated PDDL matches the natural language, set "accepted": true.\\   - if "accepted" is false, every listed issue must describe a real change that should\\     be made to objects, :init, or :goal.\\   - do not say that the problem is correct or fully aligned while also rejecting it\\   - if you cannot name a concrete fix, prefer acceptance.
\end{quote}

\section{Structural Matching Example}
\label{app:structural-example}

As an example, consider the following reference problem instance expressed in classical PDDL:
\begin{verbatim}
	(:objects a b c)
	(:init
	(on a b)
	(ontable b)
	(ontable c)
	(clear a)
	(clear c)
	)
	(:goal (on b c))
\end{verbatim}
Consider also the following automatically generated PDDL problem instance:
\begin{verbatim}
	(:objects x y z)
	(:init
	(on y z)
	(ontable z)
	(ontable x)
	(clear x)
	(clear y)
	)
	(:goal (on z x))
\end{verbatim}

A literal exact-match comparison would classify the two problems as different, since the object names do not match. Under our renaming-invariant structural criterion, the two problems are structurally identical up to a consistent renaming of objects. In the reference problem, object \texttt{a} is the clear block on top of another block, \texttt{b} is the supporting block on the table, and \texttt{c} is the other clear block on the table. The generated problem has the same relational pattern, with \texttt{y}, \texttt{z} and \texttt{x} playing these respective roles. This yields the bijection \texttt{y} $\mapsto$ \texttt{a}, \texttt{z} $\mapsto$ \texttt{b} and \texttt{x} $\mapsto$ \texttt{c}. After applying this renaming, and if one considers the predicates within \texttt{:init} as sets that are order independent, the generated initial state and goal become identical to those of the reference problem. Under this structural criterion, these two problems are therefore counted as an equivalence match.

\section{Manual Audit Examples}
\label{app:manual-audit-examples}

Table~\ref{tab:critic-manual-audit} reports the results (i.e., counts for the number of cases) for a manual audit of the LLM critic on 24 stratified examples for the \texttt{gpt-4.1-mini} run.

\begin{table}[hbp!]
	\centering
	\small
	\begin{tabular}{lcc}
		\hline
		Critic verdict & Manual faithful & Manual unfaithful \\
		\hline
		Accept & 9 & 3 \\
		Reject & 8 & 4 \\
		\hline
	\end{tabular}
	\caption{Results for a manual audit of the LLM critic on 24 stratified  examples for the \texttt{gpt-4.1-mini} run.}
	\label{tab:critic-manual-audit}
\end{table}

The following examples illustrate typical cases found in the manual critic audit. We show only the relevant fragments of the generated problem.

\paragraph{Accepted Instance with an Omitted Metric.}
In this PDDL~2.1 example from the \texttt{rover} domain, the critic accepted the generated problem, but the manual audit judged it unfaithful because the natural language description explicitly requested minimization of the recharge counter.

\begin{quote}\small
	\textbf{Natural Language Description.}
	The soil location is the far waypoint \texttt{wp2}, not the intermediate marker. Traverse through \texttt{wp1} as needed, take the soil sample at \texttt{wp2}, and communicate it to \texttt{lander0}. Track rover energy and minimize the recharge counter.
\end{quote}

\begin{small}
	\begin{verbatim}
		(:init
		(at rover0 wp0)
		(can_traverse rover0 wp0 wp1)
		(can_traverse rover0 wp1 wp2)
		(at_soil_sample wp2)
		(equipped_for_soil_analysis rover0)
		(= (energy rover0) 100)
		(= (recharges) 0)
		(at_lander lander0 wp0)
		(channel_free lander0)
		)
		(:goal (and
		(communicated_soil_data wp2)
		))
	\end{verbatim}
\end{small}

The generated instance is executable and captures the sampling goal, but it omits the requested optimization objective, e.g. a metric minimizing \texttt{recharges}. This is a false positive for the critic: operational checks can succeed while a metric-sensitive part of the NL description is lost.

\paragraph{Rejected Instance with Faithful Domain Bookkeeping.}
An example from the \texttt{depots} domain shows the opposite failure mode. The critic rejected the instance, although the manual audit judged it faithful to the natural language description, matching the curated reference encoding.

\begin{quote}\small
	\textbf{Natural Language Description.}
	The initial state states that \texttt{crate\_0} is at \texttt{depot\_1}, is clear, and is on \texttt{crate\_2}; \texttt{crate\_2} is at \texttt{depot\_1} and on \texttt{pallet\_0}. The goal is to have \texttt{crate\_0} on \texttt{pallet\_3} and \texttt{crate\_2} on \texttt{crate\_0}.
\end{quote}

\begin{small}
	\begin{verbatim}
		(:init
		(at crate_0 depot_1)
		(clear crate_0)
		(on crate_0 crate_2)
		(at crate_2 depot_1)
		(on crate_2 pallet_0)
		...
		)
		(:goal (and
		(on crate_0 pallet_3)
		(on crate_2 crate_0)
		))
	\end{verbatim}
\end{small}

In this case, the critic reported a contradiction between \texttt{(clear crate\_0)} and \texttt{(on crate\_0 crate\_2)}, although this is not a contradiction in the \texttt{depot} encoding: a crate can be clear while resting on another crate, since \texttt{clear} means that no object is on top of it. This illustrates a false negative caused by an over-strict interpretation of domain bookkeeping predicates.

\end{document}